\documentclass[11pt]{article}
\usepackage[preprint]{acl}

\usepackage{booktabs}
\usepackage{pifont}

\usepackage{amsmath}
\usepackage{amssymb}
\usepackage{times}

\usepackage[T1]{fontenc}
\usepackage[utf8]{inputenc}
\usepackage{microtype}
\usepackage{inconsolata}
\usepackage{graphicx}
\usepackage{float}
\usepackage{tabularx}

\title{MathShikkha: A Controlled Study of Answer-Only and Chain-of-Thought Supervision for Bangla Mathematical Reasoning in Small Language Models}

\author{
Rahma Simin Ali \\
Independent Researcher \\
\texttt{rahmasimin@gmail.com}
\And
Jawad Hossain \\
University at Albany \\
\texttt{jhossain2@albany.edu}
}

\begin{document}
\maketitle

\begin{abstract}
Mathematical reasoning remains challenging in low-resource languages such as Bangla. We study whether teacher-generated Bangla Chain-of-Thought (CoT) supervision provides benefits beyond ordinary supervised fine-tuning. We construct \textsc{MathShikkha}, a Bangla mathematical reasoning dataset with GPT-5.4-generated rationales, and fine-tune four 4B--7B student models under a matched protocol in which answer-only and CoT conditions share data splits, response-only loss masking, decoding, and scoring, differing only in the training target. In-domain, CoT provides no significant improvement over answer-only fine-tuning for three stronger backbones (paired bootstrap 95\% CIs include zero; exact McNemar $p \geq 0.17$), despite generating 15--52$\times$ more tokens, but significantly improves the weaker 4B model by 18.56 points ($p < 0.0001$). On the larger, contamination-audited BanglaMATH benchmark, this pattern reverses: CoT significantly outperforms answer-only supervision for all four models by 20.1--28.1 points (all $p < 0.0001$). Answer-only fine-tuning also reduces out-of-domain accuracy below the base model for three models, whereas CoT preserves or improves it for all four. A human study with two co-author annotators, external-expert adjudication, and Cohen's $\kappa = 0.76$--$1.00$ finds no significant CoT improvement over the base model on reasoning-content criteria; instead, its measurable effect is target-language adherence and producing inspectable reasoning. Overall, rationale supervision's value depends on backbone capability and distribution shift: in this setting, its main benefits are Bangla adherence, auditable reasoning, and out-of-domain robustness rather than improved in-domain reasoning validity.
\end{abstract}

\section{Introduction}
\begin{figure*}[t]
\centering
\includegraphics[width=\textwidth]{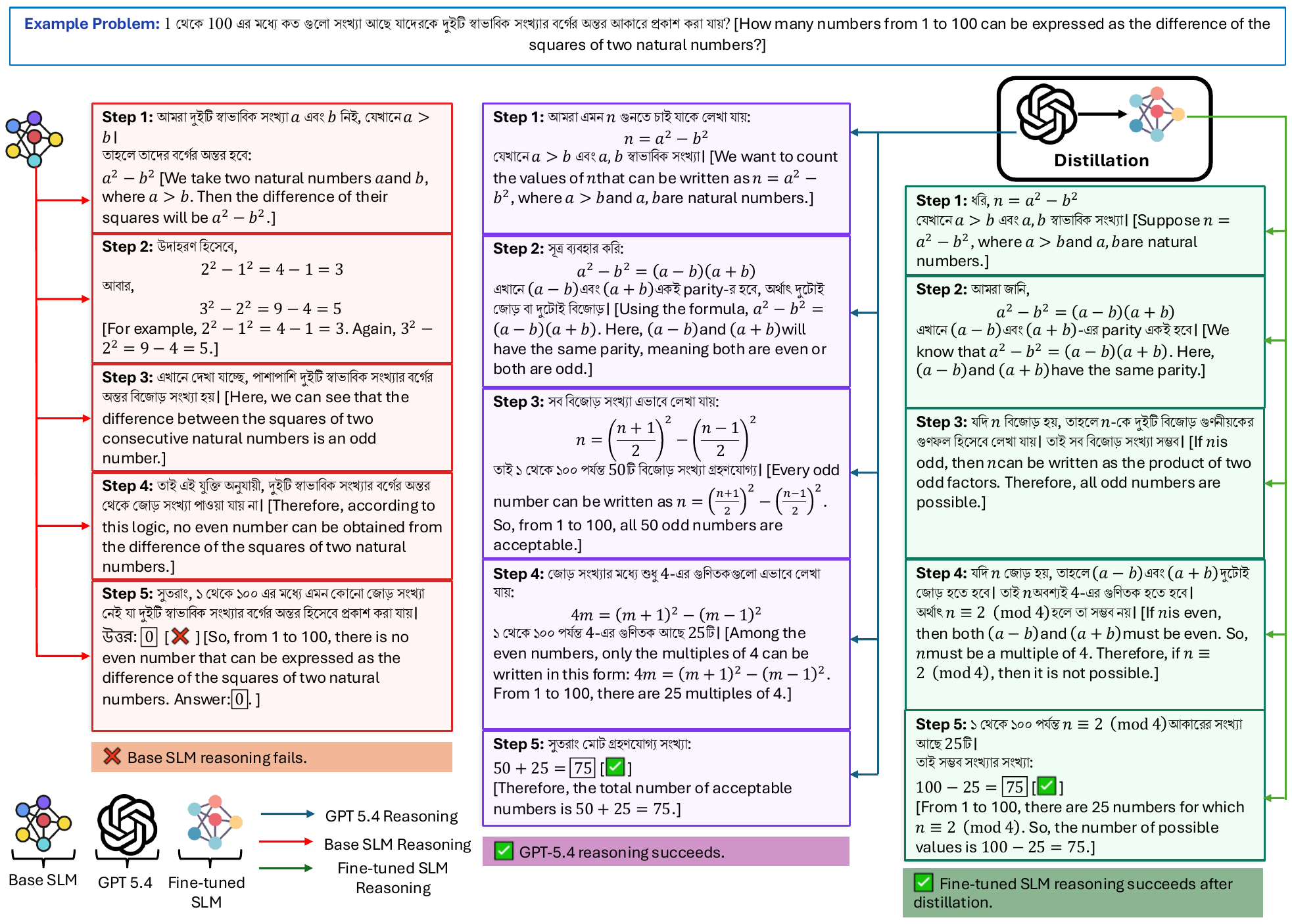}
\caption{Example reasoning comparison on a Bangla math problem. The Base SLM fails to solve the problem, whereas GPT-5.4 solves it correctly using factorization and parity analysis. After distillation, the fine-tuned SLM also reaches the correct answer, demonstrating successful transfer of reasoning ability from GPT-5.4 to the SLM.}
\label{fig:abstract_method}
\end{figure*}

Mathematical reasoning is a central capability for educational AI systems, tutoring agents, and scientific problem solving. Recent large language models (LLMs) have shown strong performance on mathematical reasoning tasks, especially when prompted or trained to produce intermediate Chain-of-Thought (CoT) rationales. However, these gains remain highly uneven across languages. In low-resource languages such as Bangla, even models with general multilingual ability often struggle to generate coherent step-by-step mathematical explanations, preserve problem constraints, and produce correct final answers.

This limitation is especially important in educational settings. For a math tutor, the final answer alone is not sufficient: students need explanations that show how quantities are related, which operations are applied, and why each step follows from the previous one. Existing Bangla-capable small language models (SLMs) frequently fail in this setting. They may skip intermediate steps, mix Bangla and English, misinterpret word-problem semantics, or produce fluent but mathematically invalid reasoning. At the same time, relying on large proprietary teacher models at inference time is costly and impractical for resource-constrained deployment.

A promising alternative is to use strong LLMs as teachers and transfer their reasoning behavior into smaller models through distillation. Prior work has shown that CoT rationales can serve as useful supervision for improving reasoning in smaller models, but the overwhelming majority of this evidence compares CoT-tuned models against models with \emph{no} fine-tuning at all, rather than against a fine-tuned model trained on the identical data with the identical objective except for the presence of the rationale. This leaves an important question unanswered: when a student model is already being fine-tuned on the target task and language, does teaching it to \emph{explain} its answer in Bangla provide any additional benefit over simply teaching it to \emph{produce} the answer?

We introduce \textsc{MathShikkha}, a Chain-of-Thought distillation framework for Bangla mathematical reasoning, and use it to answer this question directly. Given Bangla math problems and gold final answers, \textsc{MathShikkha} uses a strong teacher LLM to generate detailed Bangla reasoning traces, which are used as supervised targets for QLoRA-based fine-tuning of Bangla-capable SLMs. Critically, we complement this with a \emph{matched} answer-only fine-tuning condition that shares the identical train/validation split, response-only loss masking implementation, decoding configuration, and answer normalization pipeline as the CoT condition, differing only in the assistant training target and generation budget. This lets us isolate the causal contribution of rationale supervision from the general effect of supervised fine-tuning on the task and language.

Our central finding is that the answer to this question is not fixed: it depends both on the backbone and, crucially, on whether the model is evaluated in- or out-of-domain. \emph{In-domain}, on our held-out test split, the matched comparison reveals that Bangla CoT supervision is not uniformly beneficial: for three of four backbones, CoT supervision is statistically indistinguishable from answer-only fine-tuning despite generating 15--52$\times$ more tokens at inference time, and it helps significantly only for the smallest and weakest backbone. \emph{Out-of-domain}, on a larger, contamination-audited slice of the BanglaMATH benchmark, this pattern reverses sharply: CoT supervision significantly outperforms matched answer-only supervision for every model we test, by 20--28 points. Moreover, answer-only fine-tuning substantially damages out-of-domain accuracy relative to the non-fine-tuned base model for three of four models, while Mathstral remains approximately unchanged; CoT supervision preserves or improves external accuracy across all four models. Together these results indicate that rationale supervision's chief practical value is not raising in-domain accuracy on already-competent backbones, but protecting against the generalization loss that answer-only fine-tuning can induce.

Unlike prior Bangla mathematical reasoning resources that primarily benchmark or collect problem-solution pairs, our work uses a controlled experimental design---matched conditions, paired significance testing, and a decontaminated external evaluation---to isolate what teacher-generated Bangla reasoning traces actually contribute beyond ordinary task and language adaptation.

Our contributions are:
\begin{enumerate}
    \item We construct a Bangla mathematical reasoning dataset with teacher-generated step-by-step rationales and gold final answers, and a matched answer-only fine-tuning protocol that isolates the causal effect of rationale supervision.
    \item We show, using paired bootstrap confidence intervals and exact McNemar tests, that in-domain the benefit of Bangla CoT supervision over matched answer-only supervision is capability-dependent: significant only for the weakest backbone, and statistically indistinguishable from zero for stronger backbones despite substantially higher inference cost.
    \item We show, on a larger contamination-audited external benchmark, that this pattern reverses: CoT supervision significantly outperforms matched answer-only supervision for every model (by 20--28 points). Answer-only fine-tuning degrades out-of-domain accuracy below the base model for three of four models, while Mathstral remains approximately unchanged; CoT supervision preserves or improves external accuracy across all four models.
    \item We conduct a human study of intermediate reasoning quality across all three conditions (eight criteria, external-expert adjudication, $\kappa = 0.76$--$1.00$) showing that answer-only supervision yields no inspectable reasoning, that CoT fine-tuning does not significantly improve any reasoning-content criterion over the base model---its only significant effect being target-language adherence---and that 15.4\% of correct answers rest on invalid reasoning, so accuracy overstates reasoning quality.
    \item We provide further quantitative analyses of token efficiency, per-category external performance, a documented contamination audit, and a comparison to a purpose-built Bangla-specific model (TigerLLM-9B-it), which our smaller distilled models match, to contextualize when and why rationale supervision is worth its cost.
\end{enumerate}

\section{Related Work}
\label{sec:related_work}

Chain-of-thought (CoT) prompting has become a standard technique for improving multi-step reasoning in large language models. Prior work shows that providing intermediate reasoning steps can substantially improve performance on arithmetic, symbolic, and commonsense reasoning tasks, especially for sufficiently large models~\cite{wei2022chain,kojima2022large,sprague2025cot}. However, CoT prompting mainly acts as an inference-time strategy and depends heavily on the underlying reasoning ability of the base model. As a result, it remains unreliable for smaller models and for low-resource languages, where models often generate incomplete, inconsistent, or language-mixed reasoning traces. In contrast, our work uses CoT as a supervision signal for transferring mathematical reasoning behavior from a strong teacher LLM into compact Bangla-capable student models, and directly tests whether this signal adds value beyond ordinary supervised fine-tuning.

A growing body of work studies reasoning distillation from large teacher models to smaller students. Methods such as Fine-tune-CoT, Distilling Step-by-Step, and Symbolic Chain-of-Thought Distillation show that teacher-generated rationales can improve the reasoning ability of smaller models while reducing reliance on large models at inference time~\cite{ho2023large,hsieh2023distilling,li2023symbolic}. Related work on math-specialized instruction tuning and data augmentation, including MAmmoTH, MetaMath, DeepSeekMath, and DeepSeek-R1, further demonstrates that high-quality reasoning data and post-training can substantially improve mathematical problem solving~\cite{yue2024mammoth,yu2024metamath,shao2024deepseekmath,guo2025deepseek}. These studies establish the value of rationale-based supervision for mathematical reasoning, but they largely focus on English or high-resource settings, and few isolate rationale supervision from the confound of supervised fine-tuning itself via a matched answer-only condition, particularly with respect to out-of-domain generalization.

Multilingual mathematical reasoning remains comparatively underexplored. Prior benchmarks show that models often experience large performance drops when solving the same mathematical problems in non-English languages~\cite{shi2022language}. More recent multilingual evaluations also identify language inconsistency and off-target generation as persistent challenges, particularly when models are expected to reason in the target language rather than switch to English~\cite{luo2025mmath}. Bangla is especially underrepresented in this area, although recent resources such as BanglaMATH, SOMADHAN, and BMWP have begun to evaluate or collect Bengali mathematical word problems~\cite{prama2025banglamath,paul2025leveraging,mondal2025bmwp}. These works highlight the difficulty of Bangla mathematical reasoning, but most focus on benchmarking, prompting, or dataset construction rather than training compact models to generate coherent Bangla reasoning, and none directly test whether rationale supervision outperforms matched answer-only supervision.

\textsc{MathShikkha} addresses this gap by studying CoT distillation for Bangla mathematical reasoning in small language models under a controlled comparison against matched answer-only supervision, both in-domain and on a decontaminated external benchmark. A more detailed discussion of related work is provided in Appendix~\ref{app:related_work}.

\section{Methodology}
\label{sec:methodology}

We propose \textsc{MathShikkha}, a Chain-of-Thought (CoT) distillation framework for improving Bangla mathematical reasoning in small language models (SLMs). Rather than assuming that rationale supervision is beneficial, we design the framework to allow a direct, matched comparison against answer-only supervision. The framework has three stages: Bangla CoT data construction, QLoRA-based supervised fine-tuning under two matched training-target conditions, and held-out evaluation.

\subsection{Task Formulation}
\label{sec:task_formulation}

Let $\mathcal{X}$ denote Bangla mathematical problems and $\mathcal{A}$ denote final answers. Given a problem $x \in \mathcal{X}$, the model generates a reasoning trace $z$ and final answer $a \in \mathcal{A}$. In the \textbf{Bangla CoT} condition, we define the target output as
\begin{equation}
    y = [z; a],
\end{equation}
where $z$ is the Bangla CoT rationale and $a$ is the final answer. In the \textbf{answer-only} condition, the target is $y = [a]$ alone. Both conditions share the identical input problem $x$, train/validation split, loss-masking implementation, and evaluation pipeline; only the assistant target and generation budget differ.

\subsection{Stage 1: Bangla CoT Data Construction}
\label{sec:data_construction}

We construct our CoT-augmented Bangla mathematical reasoning dataset from the Bangla-Math dataset\footnote{https://huggingface.co/datasets/kawchar85/Bangla-Math}, which contains Bangla mathematical problems with gold answers. Each raw example consists of a Bangla problem $x_i$ and its corresponding final answer $a_i$. To create step-by-step reasoning supervision, we use GPT-5.4 as a teacher model to generate Bangla Chain-of-Thought solutions. The teacher is instructed to write the reasoning in Bangla, show intermediate mathematical operations explicitly, and end with a final answer consistent with the gold label.

During dataset construction, we filter examples whose teacher outputs are incomplete or do not provide a usable full reasoning trace, and we also remove duplicate problem statements to avoid repeated examples and overlap across splits. Starting from 1,455 raw examples, these filtering steps exclude 19 problems, resulting in 1,436 unique examples. The resulting CoT dataset is defined as
\begin{equation}
    \mathcal{D}_{\text{cot}}
    =
    \left\{
    (x_i, z_i, a_i)
    \right\}_{i=1}^{N},
\end{equation}
where $x_i$ is the Bangla problem, $z_i$ is the GPT-5.4-generated Bangla reasoning trace, and $a_i$ is the gold final answer.

\subsection{Dataset Statistics}
\label{sec:dataset_statistics}

Table~\ref{tab:dataset_statistics} presents the statistics of our final Bangla mathematical reasoning corpus. After filtering incomplete teacher generations and removing duplicate problem statements from the original 1,455-example corpus, the dataset contains 1,436 unique examples. We split the corpus into 1,030 training examples, 115 validation examples, and 291 held-out test examples. \textbf{This split is fixed once and loaded, rather than regenerated, for every training condition reported in this paper}, guaranteeing that all reported comparisons share an identical train/validation partition.

\begin{table}[t]
\centering
\small
\caption{Dataset statistics for the Bangla mathematical reasoning corpus after duplicate removal. Average problem and CoT lengths are computed using whitespace-level word tokenization.}
\label{tab:dataset_statistics}
\resizebox{\columnwidth}{!} {\begin{tabular}{lccc}
\toprule
\textbf{Split} & \textbf{\# Examples} & \textbf{Avg. Problem Length} & \textbf{Avg. CoT Length} \\
\midrule
Whole & 1436 & 25.98 & 53.77 \\
Train & 1030 & 25.99 & 54.22  \\
Validation & 115 & 26.43 & 51.52  \\
Test & 291 & 25.79 & 53.09 \\
\bottomrule
\end{tabular}}
\end{table}

\subsection{Teacher CoT Generation and Verification}
\label{sec:human_verification}

For each problem, GPT-5.4 generates a complete Bangla CoT solution. We verify the teacher outputs before using them for student fine-tuning. First, we extract the final answer from each teacher-generated CoT and compare it with the gold answer after normalization; outputs with incorrect final answers are discarded or regenerated. Second, we manually inspect a subset of teacher rationales for mathematical coherence, Bangla language consistency, and faithfulness to the problem statement. Minor formatting or language-mixing issues are corrected, while invalid rationales are filtered or regenerated. Since exhaustive proof-level verification is costly, we treat the teacher CoTs as high-quality distillation supervision rather than formally verified proofs.

\subsection{Stage 2: Matched Supervised Fine-Tuning}
\label{sec:cot_distillation}

For each problem $x_i$, we train two conditions on the identical example set: the \textbf{answer-only} condition, whose target is $a_i$ alone, and the \textbf{Bangla CoT} condition, whose target is $y_i = [z_i; a_i]$. The student model maximizes the autoregressive likelihood of its respective target sequence:
\begin{equation}
    p_{\theta}(y_i \mid x_i)
    =
    \prod_{t=1}^{|y_i|}
    p_{\theta}
    \left(
    y_{i,t}
    \mid
    x_i, y_{i,<t}
    \right).
\end{equation}
The training objective minimizes the negative log-likelihood:
\begin{equation}
    \mathcal{L}
    =
    -
    \frac{1}{N}
    \sum_{i=1}^{N}
    \sum_{t=1}^{|y_i|}
    \log
    p_{\theta}
    \left(
    y_{i,t}
    \mid
    x_i, y_{i,<t}
    \right).
\end{equation}

\subsection{Instruction Formatting and Loss Masking}
\label{sec:loss_masking}

Each training instance is formatted as a conversational instruction-tuning example. The prompt contains the task instruction and Bangla problem, while the assistant response contains the training target for the relevant condition. Let $u_i$ be the prompt and $y_i$ the assistant response. The serialized sequence is
\begin{equation}
    s_i = [u_i; y_i].
\end{equation}
We apply loss only to assistant response tokens and mask out prompt tokens:
\begin{equation}
    \mathcal{L}_{\text{resp}}
    =
    -
    \frac{1}{N}
    \sum_{i=1}^{N}
    \sum_{t=1}^{|s_i|}
    m_{i,t}
    \log
    p_{\theta}
    \left(
    s_{i,t}
    \mid
    s_{i,<t}
    \right),
\end{equation}
where $m_{i,t}=1$ for assistant response tokens and $m_{i,t}=0$ for prompt tokens. \textbf{We use the identical response-only masking implementation for both the answer-only and Bangla CoT conditions.} This is a deliberate methodological choice: applying full-sequence loss to one condition and response-only loss to the other would confound the effect of supervision type with the effect of loss masking. Holding masking identical ensures the two conditions differ only in the assistant training target.

\subsection{Stage 3: Parameter-Efficient Fine-Tuning}
\label{sec:qlora}

We fine-tune student models using QLoRA to reduce memory and compute requirements. The pretrained weights are quantized and frozen, while only low-rank adapter parameters are updated. For a linear layer with pretrained weight matrix $W_0$, the adapted weight is
\begin{equation}
    W_{\text{eff}}
    =
    Q(W_0)
    +
    \frac{\alpha}{r}BA,
\end{equation}
where $Q(W_0)$ is the quantized frozen weight, $A$ and $B$ are trainable low-rank matrices, $r$ is the LoRA rank, and $\alpha$ is the scaling factor.

We fine-tune four student SLMs: Qwen2.5-Math-7B-Instruct, DeepSeek-R1-Distill-Qwen-7B, Mathstral-7B-v0.1, and GanitLLM-4B\_SFT\_CGRPO. The validation split is used for checkpoint monitoring and model selection, while the held-out test set is used only for final evaluation.

\subsection{Inference and Evaluation Protocol}
\label{sec:inference}

At inference time, the fine-tuned student model receives only a Bangla mathematical problem and generates a prediction $\hat{y}_j$ (either $\hat{a}_j$ alone, or $[\hat{z}_j; \hat{a}_j]$, depending on condition). \textbf{Both conditions use identical greedy decoding} and the identical answer-extraction and normalization pipeline, avoiding the confound in which one condition is scored under stochastic sampling and the other under greedy decoding. The primary metric is final-answer accuracy:
\begin{equation}
    \text{Acc}
    =
    \frac{1}{M}
    \sum_{j=1}^{M}
    \mathbb{I}
    \left[
    \operatorname{norm}(\hat{a}_j)
    =
    \operatorname{norm}(a_j)
    \right],
\end{equation}
where $\operatorname{norm}(\cdot)$ handles superficial formatting differences such as Bangla and English numerals, whitespace, punctuation, and equivalent answer forms, \textbf{including preservation of decimal points}. All extracted and gold answers are compared as normalized strings to avoid numeric type-coercion artifacts. We additionally report paired bootstrap 95\% confidence intervals on the accuracy delta between conditions (10{,}000 resamples, paired by test-item index) and exact McNemar tests on the resulting $2\times2$ discordant-pair contingency table, since both conditions are evaluated on the identical held-out items.

\section{Results and Analysis}
\label{sec:results}

We organize our results around the central causal question and its settings. \textbf{RQ1: In-domain, does Bangla CoT supervision outperform matched answer-only supervision?} (Section~\ref{sec:matched_comparison}). \textbf{RQ2: Does the same relationship hold on a larger, decontaminated external benchmark?} (Section~\ref{sec:external}). \textbf{RQ3: When accuracy improves, is it because the intermediate reasoning is more valid, or mainly because the model stays on-target in Bangla?} (Section~\ref{sec:reasoning_quality}). We then situate these against prompting-only baselines (Section~\ref{sec:main_results}), a Bangla-specific baseline (Section~\ref{sec:tiger_baseline}), token efficiency (Section~\ref{sec:token_efficiency}), and training dynamics (Section~\ref{sec:training_dynamics}).

\subsection{In-Domain: Matched Answer-Only vs.\ Bangla CoT (RQ1)}
\label{sec:matched_comparison}

Table~\ref{tab:matched_comparison} reports final-answer accuracy for the answer-only and Bangla CoT conditions on the held-out test set under the matched protocol described in Section~\ref{sec:methodology}, with paired bootstrap 95\% confidence intervals and exact McNemar tests on the accuracy delta.

\begin{table*}[t]
\centering
\small
\caption{In-domain matched comparison of Answer-Only and Bangla CoT supervised fine-tuning on the held-out test set. Both conditions share an identical train/validation split, response-only loss masking, greedy decoding, and answer normalization pipeline, differing only in the assistant training target. $\Delta$ is CoT $-$ Answer-Only, in percentage points. 95\% CIs are paired bootstrap (10{,}000 resamples, same resampled indices applied to both conditions). McNemar tests are exact, computed on the discordant pairs of the paired 291-item test set.}
\label{tab:matched_comparison}
\resizebox{\textwidth}{!}{%
\begin{tabular}{lccccccc}
\toprule
\textbf{Model} & \textbf{AO (\%)} & \textbf{CoT (\%)} & \textbf{$\Delta$ (pts)} & \textbf{95\% CI} & \textbf{Discordant $n$} & \textbf{McNemar $p$} & \textbf{Sig.\ ($\alpha{=}0.05$)} \\
\midrule
Qwen2.5-Math-7B-Instruct & 49.48 & 47.42 & $-2.06$ & [$-8.59$, $4.81$] & 100 & 0.6173 & No \\
DeepSeek-R1-Distill-Qwen-7B & 47.08 & 43.99 & $-3.09$ & [$-10.31$, $4.12$] & 111 & 0.4478 & No \\
Mathstral-7B-v0.1 & 48.80 & 43.64 & $-5.15$ & [$-12.03$, $1.72$] & 105 & 0.1716 & No \\
GanitLLM-4B\_SFT\_CGRPO & 29.55 & 48.11 & $+18.56$ & [$12.03$, $24.74$] & 100 & $<0.0001$ & \textbf{Yes} \\
\bottomrule
\end{tabular}%
}
\end{table*}

The pattern is striking and consistent across three of the four backbones. For Qwen2.5-Math-7B-Instruct, DeepSeek-R1-Distill-Qwen-7B, and Mathstral-7B-v0.1---all already reasonably strong answer-only performers (49.48\%, 47.08\%, and 48.80\%)---adding Bangla CoT supervision does not yield a significant improvement: the paired bootstrap confidence intervals on the delta all include zero, and exact McNemar tests are far from significance ($p=0.6173$, $0.4478$, and $0.1716$ respectively). The point estimates are numerically negative for all three ($-2.06$, $-3.09$, and $-5.15$ points), i.e.\ CoT supervision is if anything slightly worse than answer-only supervision in-domain for these backbones.

In sharp contrast, GanitLLM-4B\_SFT\_CGRPO---the smallest backbone and the weakest answer-only performer (29.55\%)---shows a large and highly significant gain (+18.56 points; 95\% CI [12.03, 24.74]; $p<0.0001$). This is the only in-domain model for which the confidence interval excludes zero.

\paragraph{Interpretation.} In-domain, the value of explicit Bangla rationale supervision is \emph{capability-dependent}. Backbones already competent under answer-only supervision (roughly 47--52\% here) do not benefit from being additionally taught to produce a step-by-step Bangla explanation; the model apparently already encodes the reasoning implicitly, and the explicit rationale adds training- and inference-time cost without a corresponding in-domain accuracy benefit. A backbone starting from a substantially weaker answer-only baseline (29.55\%), however, benefits from the scaffolding rationale supervision provides. As Section~\ref{sec:external} shows, this in-domain reading is incomplete on its own: the picture changes once we test generalization.

\paragraph{Seed robustness.} To verify that this in-domain reading is not an artifact of a single training run, we repeat the matched Answer-Only vs.\ Bangla CoT comparison for Qwen2.5-Math-7B-Instruct under a second random seed, using the identical protocol (Table~\ref{tab:seed_robustness}). The qualitative conclusion is stable across seeds: in \emph{both} runs, Bangla CoT supervision fails to outperform matched answer-only supervision in-domain, and the sign of the effect is negative in both cases ($\Delta = -2.06$ and $-10.99$ points). Neither seed shows CoT significantly \emph{improving} over answer-only; under seed 123, answer-only supervision in fact significantly \emph{outperforms} CoT ($p = 0.0001$). The \emph{magnitude} of the gap varies non-trivially across seeds ($-2.1$ to $-11.0$ points), which we report transparently rather than average away; its \emph{direction}, however, never favors CoT. This reinforces our central in-domain finding---that explicit Bangla rationale supervision buys no in-domain accuracy for an already-strong backbone---and is orthogonal to the out-of-domain (Section~\ref{sec:external}) and reasoning-quality (Section~\ref{sec:reasoning_quality}) results, which are computed on separate evaluations.

\begin{table}[t]
\centering
\small
\setlength{\tabcolsep}{3.5pt}
\caption{Seed robustness of the in-domain matched comparison for Qwen2.5-Math-7B-Instruct. Both seeds use the identical protocol (shared split, response-only masking, greedy decoding, and type-safe scoring) on the same 291-item test set. $\Delta$ is CoT $-$ Answer-Only in percentage points; McNemar tests are exact on the paired discordant items. In both seeds CoT fails to outperform answer-only supervision; under seed 123 answer-only significantly wins.}
\label{tab:seed_robustness}
\begin{tabular}{lcccc}
\toprule
\textbf{Seed} & \textbf{AO (\%)} & \textbf{CoT (\%)} & \textbf{$\Delta$ (pts)} & \textbf{McNemar $p$} \\
\midrule
42  & 49.48 & 47.42 & $-2.06$  & 0.6173 \\
123 & 52.23 & 41.24 & $-10.99$ & 0.0001 \\
\bottomrule
\end{tabular}
\end{table}

\paragraph{Methodological note.} Isolating the effect of supervision type requires that the two conditions differ \emph{only} in the training target. We therefore hold constant, across the answer-only and CoT conditions, the loss-masking implementation (response-only loss for both), the train/validation split (loaded once and shared, never regenerated per condition), the answer-normalization function (which preserves decimal points and compares extracted and gold answers as normalized strings to avoid numeric type-coercion artifacts), and the decoding strategy (greedy for both). All accuracy numbers in this paper are produced by a single, type-safe scoring pipeline applied uniformly across models and conditions. Comparisons that do not control these factors---for example, using full-sequence loss for one condition and response-only loss for the other, or stochastic decoding for one and greedy for the other---confound supervision type with training and scoring differences and can reverse the apparent sign of the effect; we avoid them.

\subsection{Out-of-Domain: External Evaluation on BanglaMATH (RQ2)}
\label{sec:external}

The in-domain result raises an obvious question: if CoT supervision buys nothing over answer-only supervision for strong backbones, is rationale supervision simply unnecessary for them? Testing generalization answers this in the negative. We evaluate every condition on a larger, contamination-audited external benchmark drawn from BanglaMATH~\cite{prama2025banglamath}, a collection of Bangladeshi school-textbook word problems entirely separate from our training corpus.

\paragraph{Contamination audit.} Establishing that external performance reflects genuine generalization requires ruling out train/test overlap. Starting from 435 BanglaMATH candidate problems, we ran a three-stage audit against all 1{,}030 training problems: (i) exact normalized-text overlap, (ii) character-level near-duplicate similarity (threshold $0.85$), and (iii) similarity after masking all numbers and variables, which catches ``same template, different numbers'' leakage (threshold $0.90$). Stages (i) and (ii) flagged zero problems. Stage (iii) flagged a single problem, which manual inspection confirmed to be a false positive: an LCM problem and a GCD problem that collapse to the same skeleton only after number masking, sharing no actual content. Across all 435 candidates, the maximum problem-level similarity to any training problem was $0.84$, with a median of $0.53$. We therefore treat the benchmark as uncontaminated; per-stage counts are given in Appendix~\ref{app:contamination}. Models are evaluated on a 432-example clean set produced by an exact-deduplication pass, which this fuller three-stage audit independently confirms is free of genuine contamination.

\paragraph{Matched external results.} Table~\ref{tab:external} reports base, answer-only, and Bangla CoT accuracy for each model. To keep the three conditions strictly comparable, each model's accuracies are computed on the common set of items for which all of its available conditions produced output (see the per-model $n$; DeepSeek and GanitLLM fall slightly below 432 due to a small number of non-generating or, for GanitLLM base, ungenerated items).

\begin{table*}[t]
\centering
\small
\caption{Out-of-domain evaluation on the contamination-audited BanglaMATH benchmark. Each model's three conditions are scored on the common set of items all its conditions generated ($n$ column). $\Delta$ is CoT $-$ Answer-Only, in percentage points; McNemar tests are exact on the paired discordant items}
\label{tab:external}
\resizebox{\textwidth}{!}{%
\begin{tabular}{lcccccc}
\toprule
\textbf{Model} & \textbf{$n$} & \textbf{Base (\%)} & \textbf{Answer-Only (\%)} & \textbf{Bangla CoT (\%)} & \textbf{$\Delta$ (CoT$-$AO)} & \textbf{McNemar $p$} \\
\midrule
Qwen2.5-Math-7B-Instruct & 432 & 68.75 & 46.53 & 66.67 & $+20.14$ & $<0.0001$ \\
DeepSeek-R1-Distill-Qwen-7B & 427 & 58.55 & 37.70 & 65.81 & $+28.10$ & $<0.0001$ \\
GanitLLM-4B\_SFT\_CGRPO & 384 & 67.19 & 45.05 & 72.14 & $+27.08$ & $<0.0001$ \\
Mathstral-7B-v0.1 & 432 & 25.69 & 26.16 & 49.77 & $+23.61$ & $<0.0001$ \\
\bottomrule
\end{tabular}%
}
\end{table*}

Two findings stand out, and both invert the naive in-domain reading.

First, \textbf{CoT supervision decisively outperforms matched answer-only supervision out-of-domain for all four models.} The gaps are large (+20.14, +28.10, +27.08, and +23.61 points for Qwen, DeepSeek, GanitLLM, and Mathstral, respectively) and all are highly significant ($p < 0.0001$). This contrasts sharply with the in-domain results for Qwen, DeepSeek, and Mathstral, where CoT supervision was statistically indistinguishable from answer-only fine-tuning.

Second, \textbf{answer-only fine-tuning degrades out-of-domain accuracy below the non-fine-tuned base model for three of four models, while Mathstral remains approximately unchanged.} Qwen falls from a 68.75\% base accuracy to 46.53\% under answer-only fine-tuning; DeepSeek from 58.55\% to 37.70\%; and GanitLLM from 67.19\% to 45.05\%. Mathstral, in contrast, changes only slightly from 25.69\% to 26.16\%. Bangla CoT fine-tuning, by contrast, preserves or improves external accuracy across all four models: Qwen remains close to its base performance (66.67\% vs.\ 68.75\%), while DeepSeek improves from 58.55\% to 65.81\%, GanitLLM from 67.19\% to 72.14\%, and Mathstral from 25.69\% to 49.77\%. A natural interpretation is that answer-only supervision can narrowly adapt the model to the in-domain answer format and problem distribution at the cost of transferable competence, whereas requiring the model to reason step-by-step in Bangla better preserves that competence. This directly addresses the concern that our gains might reflect ordinary supervised fine-tuning rather than reasoning: out-of-domain, rationale supervision protects against the generalization loss that plain answer supervision can induce.

\paragraph{Performance by category and difficulty.} BanglaMATH provides grade level (6--8), reasoning-step count, and answer digit-length metadata, letting us examine where the CoT advantage concentrates. Table~\ref{tab:external_breakdown} summarizes the Qwen and DeepSeek CoT-vs-answer-only gap by reasoning-step count. For both models the answer-only condition degrades sharply as problems require more reasoning steps (e.g.\ Qwen answer-only falls from 56.6\% on 1-step problems to 16.7\% on 4-step problems), while the CoT condition degrades far more gracefully---consistent with the interpretation that rationale supervision preserves multi-step competence that answer-only supervision discards. Full per-grade, per-step, and per-digit tables for all models are provided in Appendix~\ref{app:external_breakdown}.

\begin{table}[t]
\centering
\small
\caption{External accuracy (\%) by reasoning-step count for the two strong backbones, answer-only (AO) vs.\ Bangla CoT. The AO$\to$CoT gap widens with reasoning depth. ($n$ per bucket in parentheses; the single 5-step item is omitted.)}
\label{tab:external_breakdown}
\resizebox{\columnwidth}{!}{%
\begin{tabular}{lcccc}
\toprule
& \multicolumn{2}{c}{\textbf{Qwen}} & \multicolumn{2}{c}{\textbf{DeepSeek}} \\
\textbf{Steps} & \textbf{AO} & \textbf{CoT} & \textbf{AO} & \textbf{CoT} \\
\midrule
1 ($n{=}145$) & 56.6 & 72.4 & 50.3 & 62.8 \\
2 ($n{=}173$) & 52.6 & 65.3 & 37.0 & 70.5 \\
3 ($n{=}101$) & 25.7 & 62.4 & 22.8 & 64.4 \\
4 ($n{=}12$)  & 16.7 & 58.3 & \phantom{0}8.3 & 66.7 \\
\bottomrule
\end{tabular}%
}
\end{table}

\subsection{Fine-Tuning vs.\ Prompting-Only Baselines}
\label{sec:main_results}

Both forms of supervised fine-tuning substantially outperform prompting-only inference on the same base models. Table~\ref{tab:main_results} reports final-answer accuracy for zero-shot direct prompting, few-shot CoT prompting, and Bangla CoT fine-tuning against each model's non-fine-tuned baseline. This comparison characterizes the general gap between prompting and fine-tuning; it is not evidence that the gain is attributable to CoT rather than to fine-tuning per se. Sections~\ref{sec:matched_comparison}--\ref{sec:external} directly address this question through matched answer-only and CoT comparisons.

Qwen2.5-Math-7B-Instruct improves from a 37.11\% non-fine-tuned baseline to 47.42\% after CoT fine-tuning; DeepSeek-R1-Distill-Qwen-7B from 28.18\% to 43.99\%; Mathstral-7B-v0.1 from 6.19\% to 43.64\%; and GanitLLM-4B\_SFT\_CGRPO from 12.37\% to 48.11\%. Few-shot CoT prompting is generally weaker than fine-tuning of either kind, indicating that in-context demonstrations alone are insufficient for robust Bangla mathematical reasoning in these models.

\begin{table*}[t]
\centering
\small
\caption{
Final-answer accuracy (\%) on the held-out test set: non-fine-tuned baselines and prompting-only settings vs.\ Bangla CoT fine-tuning. $\Delta$ is the improvement of CoT fine-tuning over the non-fine-tuned baseline. This table characterizes the prompting-vs-fine-tuning gap generally; see Tables~\ref{tab:matched_comparison} and~\ref{tab:external} for the causal answer-only-vs-CoT comparison.
}
\label{tab:main_results}
\resizebox{\textwidth}{!}{%
\begin{tabular}{lcccccc}
\toprule
\textbf{Model} & \textbf{Params} & \textbf{Zero-Shot Direct Prompt} & \textbf{Few-Shot CoT} & \textbf{Non-FT Baseline} & \textbf{Bangla CoT} & \textbf{$\Delta$} \\
\midrule
Qwen2.5-Math-7B-Instruct 
& 7B & 5.84\% & 17.53\% & 37.11\% & \textbf{47.42\%} & +10.31 \\

DeepSeek-R1-Distill-Qwen-7B 
& 7B & 6.19\% & 30.58\% & 28.18\% & \textbf{43.99\%} & +15.81 \\

Mathstral-7B-v0.1 
& 7B & 11.34\% & 8.93\% & 6.19\% & \textbf{43.64\%} & +37.45 \\

GanitLLM-4B\_SFT\_CGRPO 
& 4B & 18.56\% & 9.28\% & 12.37\% & \textbf{48.11\%} & +35.74 \\
\bottomrule
\end{tabular}%
}
\end{table*}

\subsection{Comparison to a Bangla-Specific Baseline}
\label{sec:tiger_baseline}

The models above are general-purpose or math-specialized LLMs that we adapt to Bangla. A natural question is whether a model \emph{purpose-built} for Bangla would outperform them, i.e.\ whether our distilled general-purpose models are genuinely competitive or merely exceed weak baselines. To answer this, we evaluate TigerLLM-9B-it, a 9B model instruction-tuned specifically for Bangla, under the identical evaluation pipeline (same 291-item internal test set, same split, masking, greedy decoding, and corrected type-safe scoring). We report it in three settings: zero-shot prompting, few-shot CoT prompting, and Bangla CoT fine-tuning on our training data using the identical protocol applied to the other models. Table~\ref{tab:tiger} reports the results.

\begin{table}[t]
\centering
\small
\caption{Bangla-specific baseline: TigerLLM-9B-it on the internal test set ($n{=}291$), under the identical evaluation pipeline as our other models. For reference, our fine-tuned models' Bangla CoT accuracy on the same test set ranges 43.64--48.11\% (Table~\ref{tab:matched_comparison}).}
\label{tab:tiger}
\resizebox{\columnwidth}{!}{%
\begin{tabular}{lcc}
\toprule
\textbf{Setting} & \textbf{Accuracy (\%)} & \textbf{$n$} \\
\midrule
Zero-shot prompting & 7.90 & 291 \\
Few-shot CoT prompting & 48.80 & 291 \\
Bangla CoT fine-tuning & 45.36 & 291 \\
\bottomrule
\end{tabular}%
}
\end{table}

Two observations follow. First, the Bangla-specific model does \emph{not} substantially outperform our distilled general-purpose models: its best configuration reaches 48.80\%, slightly above the best Bangla CoT-fine-tuned model in our 4B--7B group (GanitLLM, 48.11\%) and broadly comparable overall. A 9B model built specifically for Bangla thus performs comparably to our substantially smaller distilled models, indicating that MathShikkha makes general-purpose models competitive with a dedicated Bangla LLM rather than merely exceeding weak baselines.
Second, and unlike every other model in our study, TigerLLM's few-shot prompting (48.80\%) slightly \emph{exceeds} its Bangla CoT fine-tuning (45.36\%). We interpret this as consistent with our capability-dependence thesis (Section~\ref{sec:matched_comparison}): a model already instruction-tuned for Bangla can produce competent Bangla reasoning from in-context demonstrations alone, so additional Bangla CoT distillation on 1{,}030 examples adds little and, here, marginally less than few-shot prompting. Just as CoT supervision helps the weakest general-purpose backbone most in-domain, it helps a strong, already-Bangla-adapted model least. We note that 4 of 291 TigerLLM fine-tuned outputs (1.4\%) were flagged as degenerate or truncated at the generation ceiling; these are scored as incorrect and do not materially affect the reported accuracy.

\subsection{Token Efficiency}
\label{sec:token_efficiency}

Table~\ref{tab:matched_tokens} reports mean generated tokens per test item under the matched in-domain conditions. CoT supervision increases inference-time generation by 16.5$\times$ for Qwen (23.2 $\to$ 382.7 tokens) and 52.5$\times$ for DeepSeek (7.1 $\to$ 372.6 tokens). In-domain, this cost buys no accuracy for these two models; out-of-domain, the same cost buys a 20--28 point improvement and protection against the below-base degradation that answer-only fine-tuning incurs. The cost-benefit of rationale supervision is thus inseparable from the deployment setting: for a system that will only ever see in-distribution problems, answer-only fine-tuning is cheaper and equally accurate on strong backbones; for one expected to generalize, the token overhead of CoT supervision is well justified.

\begin{table}[t]
\centering
\small
\caption{Mean generated tokens per test item ($n{=}291$, in-domain) under the matched Answer-Only (AO) and Bangla CoT conditions, greedy decoding.}
\label{tab:matched_tokens}
\resizebox{\columnwidth}{!}{%
\begin{tabular}{llccc}
\toprule
\textbf{Model} & \textbf{Cond.} & \textbf{Mean} & \textbf{Median} & \textbf{Std} \\
\midrule
Qwen2.5-Math-7B-Instruct & AO & 23.2 & 23.0 & 1.4 \\
Qwen2.5-Math-7B-Instruct & CoT & 382.7 & 302.0 & 283.6 \\
DeepSeek-R1-Distill-Qwen-7B & AO & 7.1 & 6.0 & 10.2 \\
DeepSeek-R1-Distill-Qwen-7B & CoT & 372.6 & 302.0 & 281.6 \\
Mathstral-7B-v0.1 & AO & 26.3 & 26.0 & 1.8 \\
Mathstral-7B-v0.1 & CoT & 387.6 & 320.0 & 268.5 \\
GanitLLM-4B\_SFT\_CGRPO & AO & 23.0 & 23.0 & 1.0 \\
GanitLLM-4B\_SFT\_CGRPO & CoT & 385.0 & 294.0 & 289.0 \\
\bottomrule
\end{tabular}}
\end{table}

% Separately, fine-tuning on teacher-generated Bangla CoT substantially shortens output relative to the non-fine-tuned base model's own (unsupervised) attempt at CoT-style generation, reducing average length by 47.7--90.2\% across models (Appendix~\ref{app:add_exp}). Base models produce sprawling, often repetitive generations when prompted for CoT without having seen well-formed Bangla rationale examples; fine-tuning teaches concise, targeted reasoning.
Separately, we observe that fine-tuning on teacher-generated Bangla CoT yields more concise outputs than the non-fine-tuned base model's own (unsupervised) attempts at CoT-style generation: base models tend to produce sprawling, often repetitive generations when prompted for CoT without having seen well-formed Bangla rationale examples, whereas the fine-tuned models generate targeted reasoning averaging roughly 370--390 tokens per problem (Table~\ref{tab:matched_tokens}).

\subsection{Training Dynamics}
\label{sec:training_dynamics}

To understand the fine-tuning behavior, we monitor training and validation losses during CoT distillation. Figure~\ref{fig:loss_curve} in Appendix~\ref{app:add_exp} shows the loss trajectory. Both losses decrease sharply early, indicating rapid adaptation to the teacher-generated Bangla CoT format. Validation loss improves until the intermediate stage of training, then rises slightly while training loss continues to fall, indicating mild overfitting to the training rationales. We select the checkpoint with the lowest validation loss for evaluation.

\subsection{Intermediate Reasoning Quality (RQ3)}
\label{sec:reasoning_quality}

Final-answer accuracy does not establish that a model's reasoning is valid: a model can reach the correct answer through flawed intermediate steps, or produce sound reasoning that ends in an extraction error. A sharper concern is whether the accuracy differences in Sections~\ref{sec:matched_comparison}--\ref{sec:external} reflect better \emph{reasoning} at all, or mainly changes in language behavior. We address this directly with a human study of intermediate reasoning quality, comparing all three conditions that define our causal question: the non-fine-tuned base model under CoT-style prompting (\textsc{base}), the answer-only fine-tuned model (\textsc{ao-sft}), and the Bangla CoT fine-tuned model (\textsc{cot-sft}).

\paragraph{Study design.} We drew stratified samples balanced across correct/incorrect final answers: 130 items spanning the base and CoT-SFT conditions, and a further 50 items for the answer-only condition, all from the 291-item internal test set. Each item was scored on eight criteria: final-answer correctness (binary), and arithmetic correctness, logical validity, problem grounding, answer consistency, completeness, Bangla fluency, and language mixing (each scored on a 0/1/2 scale where applicable). Bangla fluency was not scored for the answer-only condition because those outputs typically contain only a number or short phrase and therefore do not provide substantive natural-language text for fluency assessment. Two annotators calibrated jointly on held-out items (excluded from reported agreement), then scored the remainder independently, without comparing item-level scores until both workbooks were submitted. Disagreements of at least one point on any criterion were adjudicated by an independent external annotator with expertise in Bangla mathematics. Inter-annotator agreement on the independently-scored base/CoT items, computed \emph{before} adjudication, was substantial to almost perfect across all eight criteria (Cohen's $\kappa = 0.760$--$1.000$; Table~\ref{tab:kappa}); agreement on the answer-only items was near-perfect (raw agreement $\geq 98\%$ on every applicable criterion), as answer-only outputs contain little reasoning content over which annotators can differ.

\begin{table}[t]
\centering
\small
\caption{Inter-annotator agreement (Cohen's $\kappa$) on the independently-scored base/CoT items, computed before external-expert adjudication.}
\label{tab:kappa}
\begin{tabular}{lc}
\toprule
\textbf{Criterion} & \textbf{Cohen's $\kappa$} \\
\midrule
Final-answer correctness & 1.000 \\
Arithmetic correctness & 0.813 \\
Logical validity & 0.845 \\
Problem grounding & 0.760 \\
Answer consistency & 0.839 \\
Completeness & 0.856 \\
Bangla fluency & 0.971 \\
Language mixing & 0.953 \\
\bottomrule
\end{tabular}
\end{table}

\paragraph{Answer-only supervision produces no inspectable reasoning; CoT and base do not differ in reasoning validity.} Table~\ref{tab:reasoning_quality} reports mean criterion scores per condition with a Kruskal--Wallis omnibus test across the three conditions and Bonferroni-corrected pairwise Mann--Whitney $U$ tests. Two facts stand out. First, the answer-only model scores at floor (mean 0.00) on every reasoning-content criterion---arithmetic correctness, logical validity, problem grounding, answer consistency, and completeness---significantly below both base and CoT-SFT ($p_{\text{bonf}} < 0.0001$). This is not a claim that answer-only ``reasons worse''; by construction the answer-only target contains no rationale, so there is no reasoning trace to evaluate. It quantifies the obvious but important point that answer-only supervision yields models whose reasoning cannot be inspected, audited, or debugged---a real cost in an educational setting even when final-answer accuracy is comparable.

Second, and more informative, \textsc{base} and \textsc{cot-sft} are statistically indistinguishable on every reasoning-content criterion under the stricter multiple-comparison correction (all $p_{\text{bonf}} \geq 0.27$): arithmetic (1.58 vs.\ 1.43), logical validity (1.00 vs.\ 0.94), problem grounding (0.86 vs.\ 1.14), answer consistency (1.54 vs.\ 1.51), and final-answer correctness (0.49 vs.\ 0.51). Bangla CoT fine-tuning does not measurably improve the logical validity of the reasoning steps over the base model.

\begin{table*}[t]
\centering
\small
\caption{Mean criterion scores (adjudicated) across the three conditions, with Kruskal--Wallis omnibus $p$ and Bonferroni-corrected pairwise Mann--Whitney $U$ $p$-values. Answer-only outputs contain no reasoning trace, so they score at floor on reasoning-content criteria by construction. Base and CoT-SFT do not differ significantly on any reasoning-content criterion. Bangla fluency is not meaningfully comparable for answer-only outputs (see text) and is therefore reported only for the base-vs-CoT comparison; ``---'' marks comparisons omitted for this reason. $^{\ast}$Completeness is significantly higher for the base than for CoT-SFT (attributed to base verbosity; see text).}
\label{tab:reasoning_quality}
\resizebox{\textwidth}{!}{%
\begin{tabular}{lccccccc}
\toprule
& \multicolumn{3}{c}{\textbf{Mean score}} & \textbf{Kruskal--Wallis} & \multicolumn{3}{c}{\textbf{Pairwise $p_{\text{bonf}}$}} \\
\textbf{Criterion} & \textbf{Base} & \textbf{AO-SFT} & \textbf{CoT-SFT} & \textbf{$p$} & \textbf{AO vs CoT} & \textbf{Base vs CoT} & \textbf{Base vs AO} \\
\midrule
Final-answer correctness & 0.49 & 0.46 & 0.51 & 0.83 & 1.000 & 1.000 & 1.000 \\
Arithmetic correctness & 1.58 & 0.00 & 1.43 & $<0.0001$ & $<0.0001$ & 1.000 & $<0.0001$ \\
Logical validity & 1.00 & 0.00 & 0.94 & $<0.0001$ & $<0.0001$ & 1.000 & $<0.0001$ \\
Problem grounding & 0.86 & 0.00 & 1.14 & $<0.0001$ & $<0.0001$ & 0.27 & $<0.0001$ \\
Answer consistency & 1.54 & 0.00 & 1.51 & $<0.0001$ & $<0.0001$ & 1.000 & $<0.0001$ \\
Completeness & 1.68 & 0.00 & 1.43 & $<0.0001$ & $<0.0001$ & 0.036$^{\ast}$ & $<0.0001$ \\
Language mixing & 0.00 & 0.00 & 1.78 & $<0.0001$ & $<0.0001$ & $<0.0001$ & 1.000 \\
Bangla fluency & 0.00 & --- & 1.68 & --- & --- & $<0.0001$ & --- \\
\bottomrule
\end{tabular}%
}
\end{table*}

\paragraph{Why Bangla fluency is excluded for answer-only.} We do not report a Bangla-fluency comparison for the answer-only condition. Answer-only outputs are typically a single number or short phrase, so there is essentially no explanatory text to judge for fluency; a two-token correct-script output trivially avoids disfluency and reaches ceiling for reasons unrelated to explanatory quality. Scoring such fragments on the same fluency scale as multi-sentence CoT explanations would not be a meaningful comparison. We therefore restrict the fluency comparison to base vs.\ CoT-SFT, where both produce full explanations. For that comparison the difference is large and significant (0.00 vs.\ 1.68, $p<0.0001$), but it is again mechanical: the base model generates off-target English rather than Bangla for 98.97\% of items (Table~\ref{tab:offtarget}), so its Bangla-fluency score is near zero by construction, whereas the fine-tuned model stays in Bangla for 91.75\%. The measurable effect of Bangla CoT fine-tuning is thus that the model \emph{reasons in the target language at all}, not that its reasoning steps become more logically valid.

\begin{table}[t]
\centering
\small
\caption{Off-target generation rate on the full 291-item internal test pool: fraction of outputs whose script is predominantly Bangla vs.\ English. Annotator-independent.}
\label{tab:offtarget}

\resizebox{\columnwidth}{!}{%
\begin{tabular}{lcc}
\toprule
\textbf{Condition} & \textbf{Bangla-dom.} & \textbf{English-dom.} \\
\midrule
Base (CoT-prompted) & 1.03\% & 98.97\% \\
Bangla CoT (fine-tuned) & 91.75\% & 8.25\% \\
\bottomrule
\end{tabular}%
}

\end{table}

\paragraph{Completeness.} Among base and CoT-SFT, the one reasoning-adjacent criterion that differs is completeness, which favors the \emph{base} model (1.68 vs.\ 1.43, $p_{\text{bonf}}=0.036$). We attribute this to verbosity rather than reasoning quality: the non-fine-tuned base model produces longer, more discursive generations (predominantly in off-target English; Table~\ref{tab:offtarget}), so it superficially includes more solution steps even when those steps do not improve final-answer accuracy or appear in the target language. Fine-tuning trades some of this length for concision and language adherence. We report the difference plainly rather than treat it as a reasoning gain for either model.

\paragraph{Right answer, wrong reasoning.} Classifying each item by the joint outcome of reasoning validity (logical validity, answer consistency, and arithmetic correctness all scored 2) and final-answer correctness makes the accuracy--reasoning gap concrete (Table~\ref{tab:outcome}). Across the base and CoT conditions, 15.4\% of items reach a \emph{correct final answer through reasoning judged invalid}, while correct reasoning with an incorrect answer is rare (1.5\%). This confirms that final-answer accuracy materially overstates reasoning quality on this task, and it justifies measuring intermediate reasoning directly rather than inferring it from accuracy.

\begin{table}[t]
\centering
\small
\caption{Joint outcome distribution (adjudicated, \% of items, base and CoT conditions) over reasoning validity and final-answer correctness. ``Incorrect reasoning, correct answer'' at 15.4\% shows that accuracy overstates reasoning quality.}
\label{tab:outcome}
\resizebox{\columnwidth}{!}{%
\begin{tabular}{lc}
\toprule
\textbf{Outcome} & \textbf{\% of items} \\
\midrule
Correct reasoning, correct answer & 34.6 \\
Incorrect reasoning, correct answer & 15.4 \\
Correct reasoning, incorrect answer & 1.5 \\
Incorrect reasoning, incorrect answer & 48.5 \\
\bottomrule
\end{tabular}%
}
\end{table}

\paragraph{Reconciling RQ1--RQ3.} Read together, the three results form a coherent picture. In-domain, CoT supervision does not raise accuracy over matched answer-only supervision for strong backbones (RQ1), and this study shows it does not raise step-level reasoning \emph{validity} over the base model either (RQ3): its measurable in-domain effect is keeping the model on-target in Bangla and producing an inspectable reasoning trace at all, which answer-only supervision does not. Yet out-of-domain, CoT supervision is precisely what better preserves generalization relative to answer-only supervision (RQ2), which substantially degrades base-model performance for three of four backbones. The value of Bangla rationale supervision in this low-resource setting is therefore not that it makes in-domain reasoning more correct---directly measured, it does not---but that it maintains target-language generation, yields auditable reasoning, and protects out-of-domain robustness. This is a more precise account than ``CoT improves reasoning,'' and it is the account our evidence supports.

\paragraph{Annotator disclosure.} The two primary annotators are co-authors of this paper; the adjudicator of disagreements is an external Bangla-mathematics expert who is not a co-author. Independence during scoring (no comparison of item-level scores before both workbooks were submitted), the inter-annotator agreement in Table~\ref{tab:kappa}, and external adjudication of all disagreements are the safeguards against merely confirmatory scoring. We note the co-author involvement as a limitation.

\section{Conclusion}
\label{sec:conclusion}

We introduced \textsc{MathShikkha} and used it to run a controlled, matched comparison of answer-only and Bangla CoT supervision---holding data splits, loss masking, decoding, and scoring identical across conditions---both in-domain and on a contamination-audited external benchmark. In-domain, CoT supervision's benefit over matched answer-only supervision is capability-dependent: significant only for the weakest backbone, and statistically indistinguishable from zero for stronger backbones despite a 15--52$\times$ inference-cost overhead. Out-of-domain, this pattern reverses: CoT supervision significantly outperforms answer-only supervision for every model (by 20--28 points). Answer-only fine-tuning degrades external accuracy below the base model for three of four backbones, while Mathstral remains approximately unchanged; CoT supervision preserves or improves external accuracy across all four models. A human study of intermediate reasoning quality further shows that, in-domain, CoT fine-tuning does not significantly improve any reasoning-content criterion over the base model; its only significant effect is keeping the model on-target in Bangla, and 15.4\% of correct answers rest on invalid reasoning. The practical value of rationale supervision, then, is less about raising in-domain accuracy or reasoning validity on already-competent backbones than about maintaining target-language generation and protecting against the out-of-domain generalization loss that plain answer supervision can induce. Methodologically, our results argue for evaluating rationale supervision against a matched answer-only baseline---and doing so out-of-domain and at the level of intermediate reasoning, not only via in-domain accuracy---rather than assuming its benefit.

\section*{Ethics Statement}
This work aims to improve Bangla mathematical reasoning for educational applications using teacher-generated Chain-of-Thought supervision, matched answer-only supervision, and small language model fine-tuning. The dataset consists of mathematical problem--answer pairs and generated reasoning traces, and does not intentionally include private, personally identifiable, or sensitive information. Since the reasoning traces are produced by a teacher LLM, they may still contain subtle mathematical or pedagogical errors despite answer normalization and manual inspection. Moreover, our human evaluation shows that correct final answers do not necessarily imply valid intermediate reasoning: 15.4\% of evaluated base and CoT outputs with correct answers contain reasoning judged invalid. Models trained with \textsc{MathShikkha} should therefore be used as assistive educational tools rather than authoritative tutors or graders without human oversight. Our finding that answer-only fine-tuning can degrade out-of-domain accuracy below the base model is directly relevant to deployment: practitioners should evaluate on genuinely held-out, distribution-shifted problems before deploying a fine-tuned Bangla math model, since strong in-domain accuracy can coexist with damaged generalization. We also recognize that Bangla varies across regions, curricula, and learner backgrounds; future deployment should include broader evaluation for linguistic coverage, fairness, reasoning reliability, and pedagogical usefulness across diverse Bangla-speaking communities.

\bibliography{custom}

\appendix

\clearpage
\section{Related Work}
\label{app:related_work}
\paragraph{Chain-of-thought reasoning.}
Chain-of-thought (CoT) prompting has become a standard approach for eliciting multi-step reasoning in large language models. \citet{wei2022chain} show that providing intermediate reasoning steps can substantially improve performance on arithmetic, symbolic, and commonsense reasoning tasks, particularly for sufficiently large models. \citet{kojima2022large} further demonstrate that models can be encouraged to produce reasoning traces in a zero-shot setting using simple instructions such as ``Let's think step by step.'' Subsequent work has analyzed when and why CoT is useful, showing that its benefits are most pronounced for tasks requiring mathematical or symbolic reasoning~\cite{sprague2025cot}. Our work extends this line of inquiry by asking not merely whether CoT supervision helps, but whether it helps \emph{beyond} what a matched answer-only supervision condition already achieves---and finds that the answer depends on the distributional gap between training and evaluation.

\paragraph{Reasoning distillation from LLMs to smaller models.}
A growing line of work studies whether reasoning behavior from large teacher models can be transferred into smaller student models. Fine-tune-CoT~\cite{ho2023large} uses teacher-generated rationales to fine-tune smaller models. Distilling Step-by-Step~\cite{hsieh2023distilling} demonstrates that label-rationale pairs from LLMs provide data-efficient supervision. Symbolic Chain-of-Thought Distillation~\cite{li2023symbolic} shows that rationale-based supervision improves smaller models on structured reasoning tasks. These works establish CoT distillation as effective, but primarily focus on English benchmarks and, to our knowledge, do not report a matched answer-only ablation with identical loss masking, splits, decoding, and scoring, nor test whether such an ablation's conclusion changes out-of-domain. Our work shows that it does. Complementary work such as HintMR~\cite{hossain2026hintmr} distills LLM-generated step-wise mathematical hints into a compact SLM hinter that guides a separate SLM solver during inference. This two-model framework shows that localized, context-aware guidance can substantially improve SLM mathematical reasoning while reducing reliance on large teacher models at inference time.

\paragraph{Math-specialized instruction tuning and data augmentation.}
MAmmoTH~\cite{yue2024mammoth} trains open-source models on MathInstruct, combining diverse problems with CoT and Program-of-Thought rationales. MetaMath~\cite{yu2024metamath} improves mathematical fine-tuning through question bootstrapping. These works show the centrality of training-data quality and structure, but focus on English and do not address whether similar supervision improves Bangla reasoning beyond matched answer-only supervision.

\paragraph{Math-specialized models and reinforcement learning for reasoning.}
DeepSeekMath~\cite{shao2024deepseekmath} continues pretraining on math-related web data and introduces Group Relative Policy Optimization. DeepSeek-R1~\cite{guo2025deepseek} shows that large reasoning models develop advanced reasoning through RL with rule-based rewards, and that their traces can be distilled into smaller models. Our work is complementary: instead of training a large reasoning model with RL, we study when a lightweight, matched supervised comparison for Bangla favors rationale supervision.

\paragraph{Multilingual mathematical reasoning.}
MGSM~\cite{shi2022language} shows performance drops when problems are presented in non-English languages. MMATH~\cite{luo2025mmath} covers ten languages and identifies off-target generation, where models answer or reason in an unintended language---closely related to our observation that CoT supervision markedly reduces language-mixing errors.

\paragraph{Bangla mathematical reasoning.}
Recent Bangla datasets and benchmarks---BanglaMATH~\cite{prama2025banglamath}, SOMADHAN~\cite{paul2025leveraging}, BMWP~\cite{mondal2025bmwp}---establish that Bangla mathematical reasoning is challenging, but focus on benchmarking, prompting, or dataset construction. Our work addresses a complementary gap: under what conditions Bangla CoT supervision actually outperforms simpler answer-only supervision, in- and out-of-domain.

\paragraph{Summary.}
Prior work shows CoT prompting helps large models and that rationale distillation transfers reasoning to smaller ones, generally evaluated against non-fine-tuned baselines. We add a matched answer-only comparison and find that the accuracy benefit of rationale supervision over answer-only supervision is capability-dependent in-domain and reverses to a consistent, significant advantage out-of-domain.

\section{Additional Experimental Setup}
\renewcommand{\thefigure}{B.\arabic{figure}}
\renewcommand{\thetable}{B.\arabic{table}}
\setcounter{figure}{0}
\setcounter{table}{0}
\label{app:add_exp}

\begin{table}[t]
\centering
\small
\caption{Hyperparameters used for QLoRA-based supervised fine-tuning (both Answer-Only and Bangla CoT conditions).}
\label{tab:hyperparameters}
\begin{tabularx}{\columnwidth}{@{}X c@{}}
\toprule
\textbf{Hyperparameter} & \textbf{Value} \\
\midrule
Number of epochs & 5 \\
Per-device train batch size & 1 \\
Per-device eval batch size & 1 \\
Gradient accumulation steps & 8 \\
Effective batch size & 8 \\
Learning rate & $2\times10^{-4}$ \\
Warmup ratio & 0.03 \\
Learning rate scheduler & Cosine \\
Optimizer & Paged AdamW 8-bit \\
Precision & bfloat16 \\
Gradient checkpointing & Enabled \\
Checkpoint saving strategy & Every 25 steps \\
Maximum saved checkpoints & 3 \\
Maximum sequence length (CoT) & 3072 \\
Packing & Disabled \\
Logging frequency & Every 5 steps \\
Logging backend & TensorBoard \\
Decoding (evaluation) & Greedy \\
\bottomrule
\end{tabularx}
\end{table}

\begin{figure}[t]
    \centering
    \includegraphics[width=0.9\linewidth]{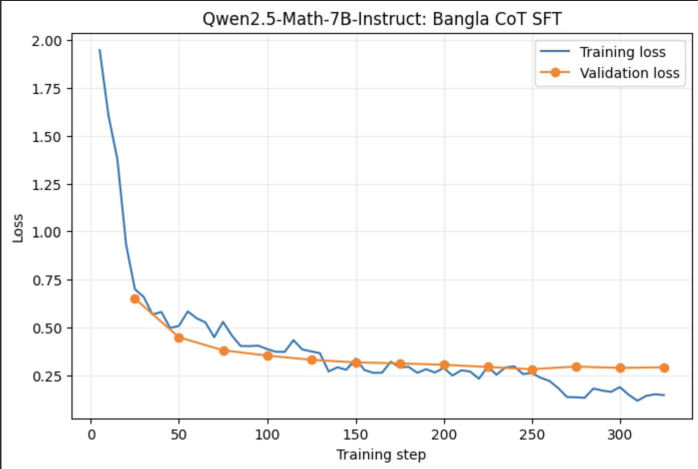}
    \caption{
    Training and validation loss during CoT fine-tuning for Qwen2.5-Math-7B-Instruct. Validation loss decreases during the early and middle stages of training, then rises slightly, indicating mild overfitting after the best checkpoint.
    }
    \label{fig:loss_curve}
\end{figure}

\section{Contamination Audit Details}
\renewcommand{\thetable}{C.\arabic{table}}
\setcounter{table}{0}
\label{app:contamination}

Table~\ref{tab:contamination_funnel} reports the three-stage contamination funnel for the BanglaMATH external candidates against the 1{,}030 training problems. All text was normalized (Bangla$\to$English numerals, whitespace and punctuation stripped, lowercased) before comparison. Near-duplicate similarity uses character-level sequence-matching ratio; masked similarity applies the same measure after replacing every digit run and Latin variable with a placeholder token, targeting ``same template, different numbers'' leakage. The single masked-similarity flag was manually inspected and found to be a false positive (an LCM problem matched against a GCD problem sharing only post-masking boilerplate), leaving zero genuine contamination. The maximum problem-level similarity to any training problem across all 435 candidates was 0.84 (mean 0.53, median 0.53).

\begin{table}[t]
\centering
\small
\caption{BanglaMATH contamination funnel against the 1{,}030 training problems.}
\label{tab:contamination_funnel}
\begin{tabular}{lc}
\toprule
\textbf{Stage} & \textbf{Count} \\
\midrule
Candidate examples & 435 \\
\quad removed --- exact overlap & 0 \\
\quad removed --- near-duplicate ($\geq 0.85$) & 0 \\
\quad removed --- masked-similarity ($\geq 0.90$) & 1$^{\dagger}$ \\
Total genuine contamination & 0 \\
\bottomrule
\end{tabular}
\end{table}

\noindent{\footnotesize $^{\dagger}$False positive on manual inspection (see text); not genuine contamination.}

\section{External Evaluation Breakdowns}
\renewcommand{\thetable}{D.\arabic{table}}
\setcounter{table}{0}
\label{app:external_breakdown}

Tables~\ref{tab:breakdown_grade}--\ref{tab:breakdown_digit} report external BanglaMATH accuracy by grade level, reasoning-step count, and answer digit-length for all models and available conditions. Empty buckets and buckets with very few examples ($n \leq 2$) are noise-dominated and should be read with caution.

\begin{table}[H]
\centering
\small
\caption{External accuracy (\%) by grade level. AO = answer-only, CoT = Bangla CoT. 
Overall benchmark counts are grade 6 = 214, grade 7 = 119, and grade 8 = 99; 
GanitLLM Base is evaluated on its 384-item available subset 
(grade 6 = 214, grade 7 = 119, grade 8 = 51).}
\label{tab:breakdown_grade}
\resizebox{\columnwidth}{!}{%
\begin{tabular}{lcccccccccccc}
\toprule
& \multicolumn{3}{c}{\textbf{Qwen}}
& \multicolumn{3}{c}{\textbf{DeepSeek}}
& \multicolumn{3}{c}{\textbf{Mathstral}}
& \multicolumn{3}{c}{\textbf{Ganit}} \\
\textbf{Grade}
& \textbf{Base} & \textbf{AO} & \textbf{CoT}
& \textbf{Base} & \textbf{AO} & \textbf{CoT}
& \textbf{Base} & \textbf{AO} & \textbf{CoT}
& \textbf{Base} & \textbf{AO} & \textbf{CoT} \\
\midrule
6 & 62.6 & 51.9 & 63.6
  & 55.6 & 38.8 & 61.2
  & 33.2 & 29.0 & 48.1
  & 68.2 & 44.4 & 73.4 \\
7 & 78.2 & 43.7 & 74.0
  & 69.8 & 35.3 & 75.6
  & 19.3 & 28.6 & 56.3
  & 68.9 & 45.4 & 72.3 \\
8 & 70.7 & 38.4 & 64.7
  & 51.5 & 36.4 & 65.7
  & 17.2 & 17.2 & 45.5
  & 58.8 & 41.4 & 65.7 \\
\bottomrule
\end{tabular}}
\end{table}
 
\begin{table}[H]
\centering
\small
\caption{External accuracy (\%) by reasoning-step count. AO = answer-only, CoT = Bangla CoT.}
\label{tab:breakdown_steps}
\resizebox{\columnwidth}{!}{%
\begin{tabular}{lcccccccccccc}
\toprule
& \multicolumn{3}{c}{\textbf{Qwen}} 
& \multicolumn{3}{c}{\textbf{DeepSeek}} 
& \multicolumn{3}{c}{\textbf{Mathstral}} 
& \multicolumn{3}{c}{\textbf{Ganit}} \\

\textbf{Steps} 
& \textbf{Base} & \textbf{AO} & \textbf{CoT} 
& \textbf{Base} & \textbf{AO} & \textbf{CoT} 
& \textbf{Base} & \textbf{AO} & \textbf{CoT} 
& \textbf{Base} & \textbf{AO} & \textbf{CoT} \\

\midrule
1 & 66.9 & 56.6 & 72.4 
  & 51.0 & 50.3 & 62.8 
  & 33.1 & 32.4 & 54.5 
  & 66.9 & 57.2 & 75.2 \\

2 & 72.3 & 52.6 & 65.3 
  & 68.2 & 37.0 & 70.5 
  & 27.2 & 29.5 & 53.8 
  & 72.9 & 46.8 & 73.4 \\

3 & 66.3 & 25.7 & 62.4 
  & 54.5 & 22.8 & 64.4 
  & 14.9 & 14.9 & 38.6 
  & 57.3 & 24.8 & 64.4 \\

4 & 66.7 & 16.7 & 58.3 
  & 50.0 & \phantom{0}8.3 & 66.7 
  & \phantom{0}8.3 & \phantom{0}0.0 & 33.3 
  & 63.6 & \phantom{0}8.3 & 58.3 \\

\bottomrule
\end{tabular}}
\end{table}

\begin{table}[H]
\centering
\small
\caption{External accuracy (\%) by answer digit-length (1--5 digits shown; higher-digit buckets have $n \leq 5$ and are omitted). AO = answer-only, CoT = Bangla CoT.}
\label{tab:breakdown_digit}
\resizebox{\columnwidth}{!}{%
\begin{tabular}{lcccccccccccc}
\toprule
& \multicolumn{3}{c}{\textbf{Qwen}}
& \multicolumn{3}{c}{\textbf{DeepSeek}}
& \multicolumn{3}{c}{\textbf{Mathstral}}
& \multicolumn{3}{c}{\textbf{Ganit}} \\

\textbf{Digits}
& \textbf{Base} & \textbf{AO} & \textbf{CoT}
& \textbf{Base} & \textbf{AO} & \textbf{CoT}
& \textbf{Base} & \textbf{AO} & \textbf{CoT}
& \textbf{Base} & \textbf{AO} & \textbf{CoT} \\
\midrule

1 & 84.2 & 62.4 & 79.2
  & 68.3 & 44.6 & 73.3
  & 49.5 & 35.6 & 67.3
  & 76.3 & 55.4 & 77.2 \\

2 & 71.7 & 44.4 & 69.0
  & 63.1 & 36.4 & 71.1
  & 20.9 & 27.3 & 49.2
  & 70.7 & 40.6 & 72.7 \\

3 & 59.3 & 42.4 & 59.3
  & 49.2 & 33.9 & 57.6
  & 15.3 & 20.3 & 40.7
  & 62.7 & 40.7 & 64.4 \\

4 & 56.3 & 25.0 & 50.0
  & 43.8 & 31.3 & 56.3
  & 6.3 & 18.8 & 31.3
  & 50.0 & 31.3 & 56.3 \\

5 & 50.0 & 42.9 & 57.1
  & 42.9 & 14.3 & 42.9
  & 7.1 & 7.1 & 42.9
  & 45.5 & 35.7 & 64.3 \\

\bottomrule
\end{tabular}}
\end{table}

\end{document}